\documentclass[journal]{IEEEtran}
\usepackage{framed}
\usepackage[ruled]{algorithm2e}
\usepackage{amssymb}
\usepackage{latexsym}
\usepackage{epsfig}
\usepackage{graphicx}
\usepackage{booktabs,multirow, multicol}
\usepackage{cite}
\usepackage{epsfig}
\usepackage{caption}
 
\usepackage{epstopdf}

\usepackage{url}
\usepackage{xcolor}
\definecolor{newcolor}{rgb}{.8,.349,.1}

\ifCLASSINFOpdf

\fi

\begin{document}

\title{A Framework for Fast Face and Eye Detection}

\author{Anjith~George,~\IEEEmembership{Member,~IEEE,}, Anirban~Dasgupta,~\IEEEmembership{Member,~IEEE,}
        and~Aurobinda~Routray,~\IEEEmembership{Member,~IEEE}
\thanks{The full version of the method described in this paper is available in : Dasgupta, Anirban, Anjith George, S. L. Happy, and Aurobinda Routray. "A Vision-Based System for Monitoring the Loss of Attention in Automotive Drivers." Intelligent Transportation Systems, IEEE Transactions on 14, no. 4 (2013): 1825-1838.}}

\maketitle

\begin{abstract}
Face detection is an essential step in many computer vision applications like surveillance, tracking, medical analysis, facial expression analysis etc. Several approaches have been made in the direction of face detection. Among them, Haar-like features based method is a robust method. In spite of the robustness, Haar - like features work with some limitations. However, with some simple modifications in the algorithm, its performance can be made faster and more robust. The present work refers to the increase in speed of operation of the original algorithm by down sampling the frames and its analysis with different scale factors. It also discusses the detection of tilted faces using an affine transformation of the input image.
\end{abstract}
\IEEEpeerreviewmaketitle

\section{Introduction}
Object detection such as face and eye detection is an important step in many vision based applications which may include video surveillance, tracking, medical analysis, facial expression analysis etc. Many researchers \cite{viola2001rapid,rowley1998neural,yang1994human,yang1996real,yang2000face,yow1997feature,zhu2000fast,jones2001rapid,rowley1998neural,ben2006fuzzy,han2000fast,hjelmaas2001face,kong2007multi,hsu2002face}  have proposed different methods for detection of face. Among these methods, the use of Haar-like features \cite{viola2001rapid,jones2001rapid,hjelmaas2001face,kong2007multi,hsu2002face} is found to be quite robust.  However, the processing rate is found to be slow for applications requiring high frame rates of processing. The classifier based on Haar like features detects frontal faces accurately. Some applications require detection of faces even with small tilts. Several approaches \cite{kong2007multi,hsu2002face,lienhart2002extended,huang2007high,mohanty2005designing} have been made to detect tilted faces.    This paper discusses the improvements on the Haar-like features based method and their analysis. There are three basic contributions in this paper. Firstly, it analyses the change of processing speed with the down sampling scale factors (SF). Secondly, it analyses the change of accuracy of detection with different SF’s. Finally, an improvement of the algorithm is done by performing an affine transformation \cite{mohanty2005designing} on the input image to detect tilted faces. The combined algorithm is implemented in a single board computer. This paper is organized as follow: Section 2 gives an introduction to face detection using Haar-like features and its modifications done to improve the robustness of the algorithm. Section 3 analyses the speed versus scale factor. In section 4, the accuracy versus scale factor is analysed. Section 5 discusses the results obtained in sections 3 and 4. Section 6 deals with the detection of tilted face using an affine transformation. Section 7 concludes the paper.
\section{The Algorithm}
\subsection{Haar-like Features}
Haar-like features \cite{viola2001rapid,jones2001rapid} are certain features in a digital image which are used in object detection. They are named so on account of their similarity with Haar wavelets. The Haar-like features are used in real-time object detection. The algorithm was first developed by Viola et. al. \cite{viola2001rapid,jones2001rapid} and was later extended by Lienhart et. al.\cite{lienhart2002extended}. The algorithm \cite{viola2001rapid} was found to achieve a 95\% accuracy rate for the detection of a human face using only 200 simple features. 
\subsection{Speeding up Operation}
In the original algorithm, the full resolution picture is examined to detect the face. Integral images \cite{viola2001rapid} are used to calculate features rapidly in multi-resolutions. Once the integral images are computed, any one of these Haar-like features can be obtained at any scale or location in constant time. From this, it is obvious that after the integral image has been calculated, the time needed for Haar-like feature based detection is constant. So in order to speed up the detection time, the time taken for the calculation of the integral images must be minimized. The time taken, to calculate the integral image, increases with the size of the image. The image is down sampled to decrease the time taken. The number of pixels in the new image is now reduced. This improves the speed of detection of face. In order to detect the eyes, a region of interest (ROI) is selected from the face region detected in the down sampled image. The ROI must be at the same resolution as the image captured from camera to detect the eyes with maximum accuracy. This is achieved by remapping ROI co-ordinates to original image. 
\subsection{Remapping ROI Co-ordinates to Original Image}
\begin{figure}[h]
\begin{center}
   \includegraphics[width=1\linewidth]{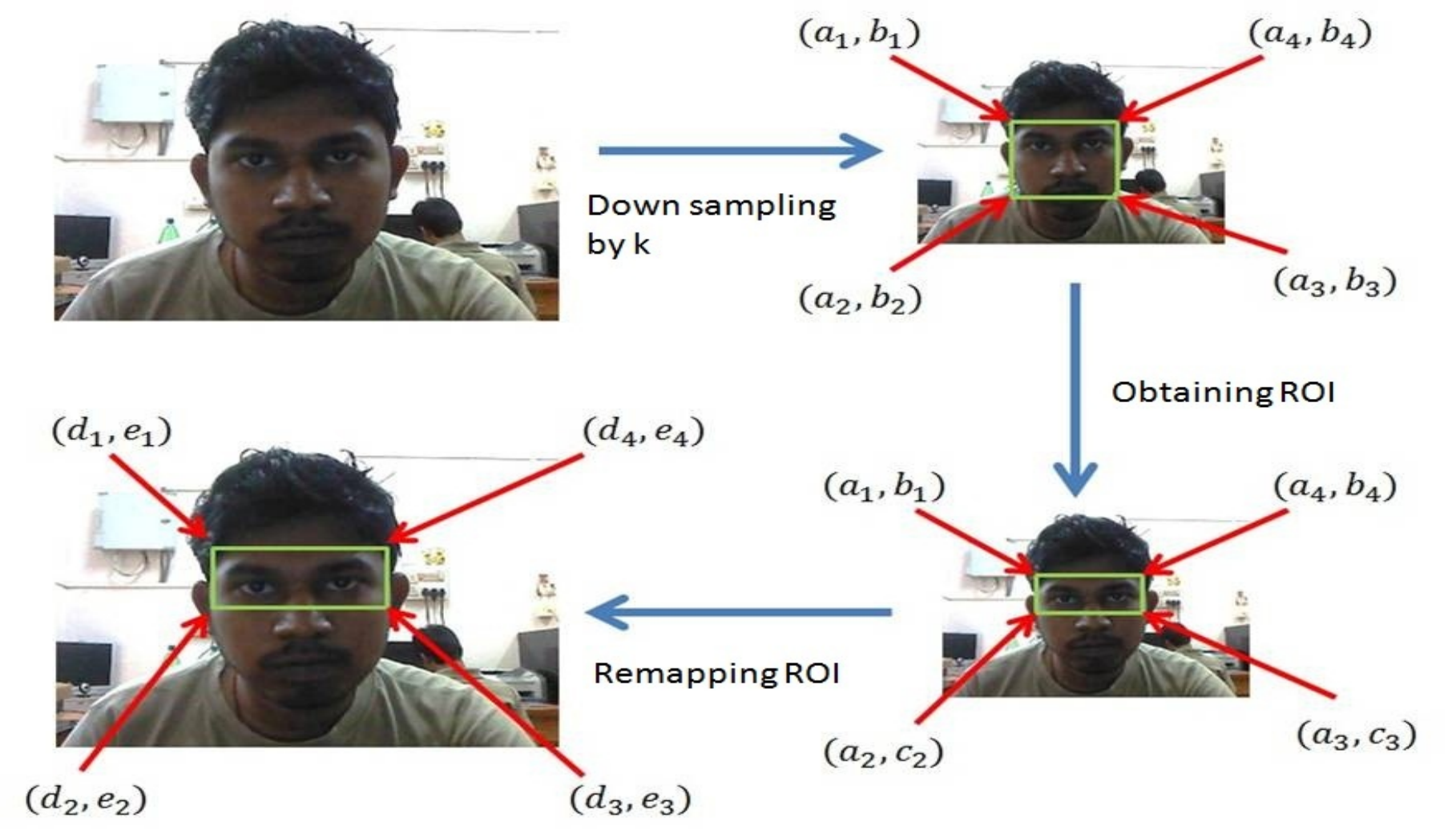}
\end{center}
   \caption{ROI remapping}
\label{fig:roi_remap}
\end{figure}

The original image is stored before it is down sampled. The coordinates of the ROI in the down sampled images are obtained, once the face detection is over. These are remapped on to the original image to obtain the ROI to detection of eyes. This method neither alters the detection rate of eyes nor the accuracy as compared to the original algorithm and at the same time, improves the processing speed up to 10 fps from 3 fps. Fig. \ref{fig:roi_remap} describes the scheme of remapping of ROI.
\section{Speed and Scale Factor Analysis}
\subsection{Experiment Design}
\begin{figure}[h]
\begin{center}
   \includegraphics[width=1\linewidth]{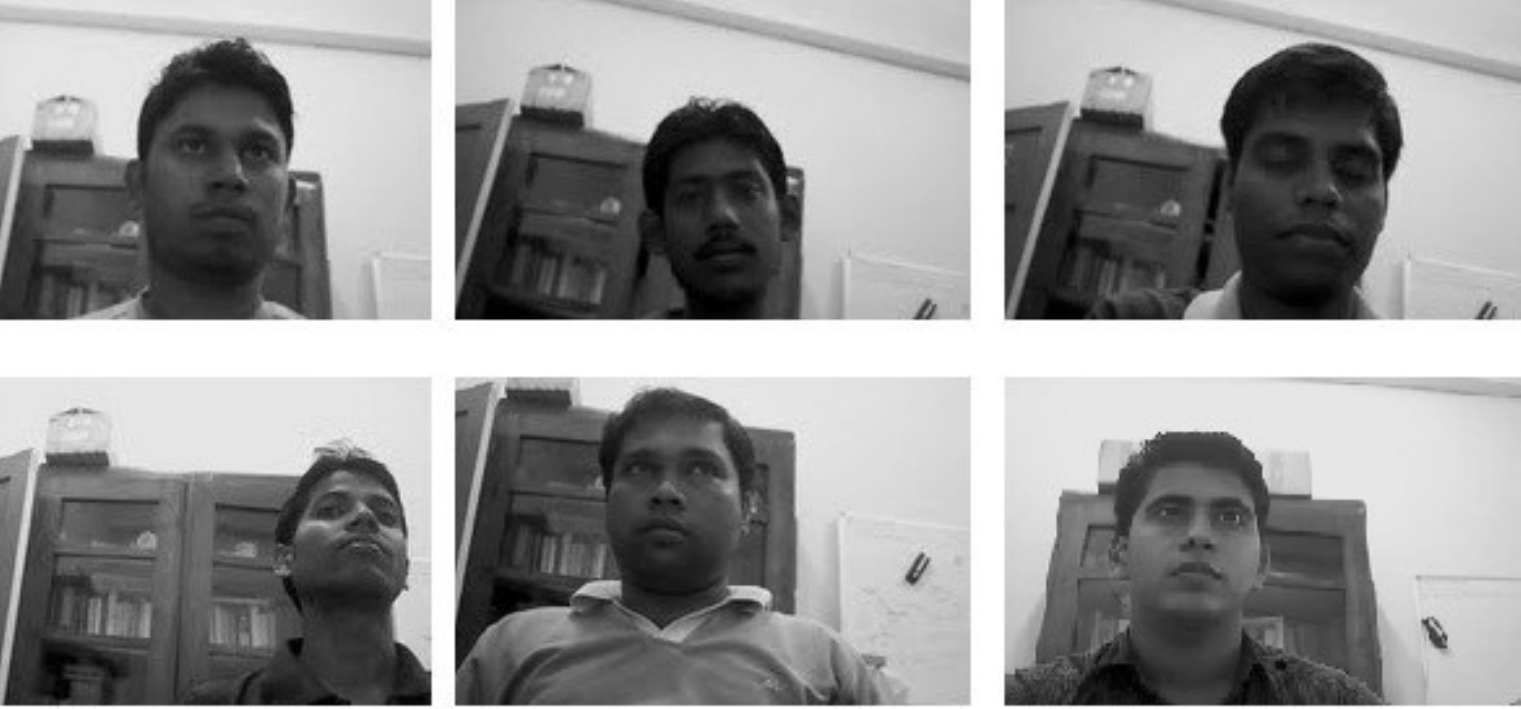}
\end{center}
   \caption{Some sample test dataset}
\label{fig:sample_image}
\end{figure}

We define the SF as follows:
\begin{equation}
SF = {{no.\,of\,vertical\,pixels\,in\,original\,frame} \over {no.\,of\,vertical\,pixels\,in\,downsampled\,frame}} 
\end{equation}
or
\begin{equation}
SF = {{no.\,of\,horizontal\,pixels\,in\,original\,frame} \over {no.\,of\,horizontal\,pixels\,in\,downsampled\,frame}}
\end{equation}
For the analysis of speed versus SF, an experiment is conducted. Six subjects are chosen and videos of facial and non facial images are recorded under laboratory conditions and stored at 30 fps at a resolution of 640 x 480 pixels in .avi format. The incoming frame is down sampled by SF’s of 2, 4, 6, 8 and 10. The processing speed for each video is noted down. The processing is done in a computer having specifications of Intel dual core processor, of speed 2.00 GHz, 2 GB of DDR2 RAM. Fig. \ref{fig:sample_image} shows some sample images extracted from the videos. 
\subsection{Results}
\begin{figure}[h]
\begin{center}
   \includegraphics[width=1\linewidth]{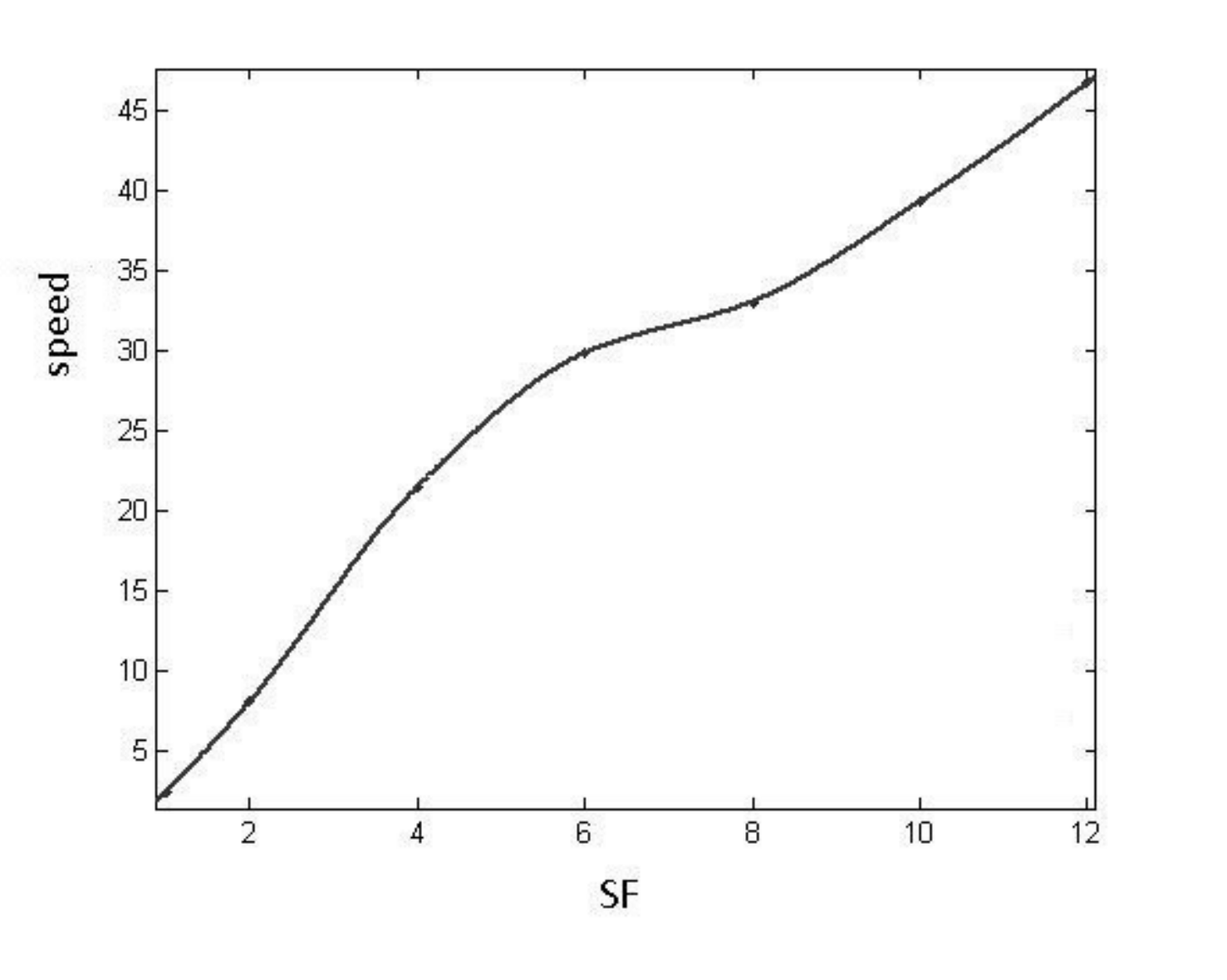}
\end{center}
   \caption{Graph showing SF vs Speed versus}
\label{fig:speed_sf}
\end{figure}

\begin{figure}[h]
\begin{center}
   \includegraphics[width=1\linewidth]{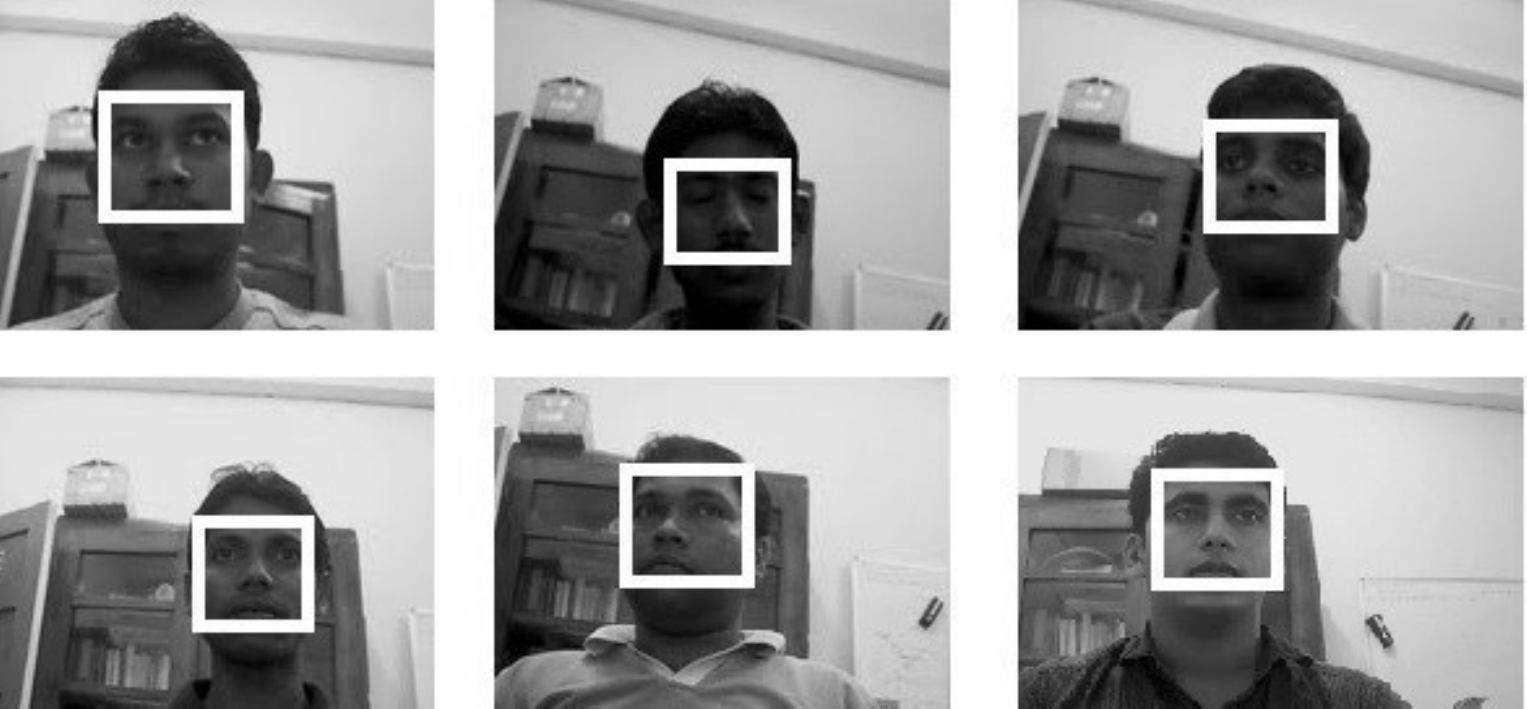}
\end{center}
   \caption{Some detection results}
\label{fig:detection_image}
\end{figure}

Face and eyes are detected as the algorithm is run on the videos. Fig. \ref{fig:detection_image} shows some face detection results. The average speed of all the subjects for each SF is plotted in Matlab against the different SF’s. Table \ref{table:speed_sf} shows the readings of speed data obtained for each subject at different SF’s.  Fig. \ref{fig:speed_sf} shows the plot of average speed versus SF.
\begin{table}[t]
\centering
\begin{tabular}{@{}llllllll@{}}
\toprule

                    & Sub1  & Sub2  & Sub3  & Sub4  & Sub5  & Sub6  & Average \\ \midrule
1                   & 2.45  & 2.13  & 2.33  & 2.43  & 2.49  & 2.6   & 2.405   \\
2                   & 9.78  & 9.09  & 7.61  & 6.26  & 8.69  & 7.1   & 8.088   \\
4                   & 17.64 & 23.75 & 20.78 & 21.96 & 21.16 & 23.58 & 21.478  \\
6                   & 28.69 & 28.39 & 31.46 & 29.1  & 29.84 & 31.43 & 29.818  \\
8                   & 33.26 & 30.63 & 35.27 & 31.45 & 35.27 & 32.26 & 33.023  \\
10                  & 40.14 & 37.54 & 43.11 & 32.33 & 43.11 & 39.55 & 39.297  \\
12                  & 46.56 & 46.56 & 46.56 & 43.11 & 46.56 & 51.08 & 46.738  \\ \bottomrule
\end{tabular}
\caption{Processing speed versus SF}
\label{table:speed_sf}
\end{table}
\section{Accuracy and Scale Factor Analysis}
\subsection{Experiment Design}
The accuracy of detection is analyzed by plotting ROC curves and then calculating the area under the curve (AUC). The same videos which were used for speed analysis are used for the analysis of accuracy vs. SF. First of all the frames are extracted as jpg image file using Free Video to JPEG Converter. Then, the images are manually marked to take into account the presence of face and eyes. A Matlab Graphical User Interface (GUI) is made to store the ground truth. Then the program for face and eye detection is run and the detection results are stored in another Excel sheet. The detection results are then compared with the ground truth to obtain the number of true positives (tp), false positives (fp), true negatives (tn) and false negatives (fn). The true positive rate (tpr) is calculated using
\begin{equation}
tpr = {{tp} \over {tp + fn}}
\end{equation}
The false positive rate (fpr) is calculated using
\begin{equation}
fpr = {{fp} \over {fp + tn}}
\label{eq:accuracy}
\end{equation}
\subsection{Results}
The plot of tpr vs fpr is called the ROC curve \cite{fawcett2004roc}. The ROC curve for each SF is obtained.\\
After obtaining the following observations, the graph of AUC versus SF is plotted using in Matlab. The accuracy \cite{fawcett2004roc} of the classifier can be calculated as in equation \ref{eq:accuracy}
\begin{equation}
Accuracy = {{tp + tn} \over {p + n}}
\end{equation}
Where $p$ is the number of positive images and $n$ the number of negative images.
\begin{table}[h]
\centering
\begin{tabular}{@{}lll@{}}
\toprule
SF & \multicolumn{1}{c}{\begin{tabular}[c]{@{}c@{}}Area Under\\  the Curve (AUC)\end{tabular}} &  \\ \midrule
1  & 0.8460                                                                                    &  \\
2  & 0.8461                                                                                    &  \\
4  & 0.8493                                                                                    &  \\
6  & 0.8274                                                                                    &  \\
8  & 0.5790                                                                                    &  \\
10 & 0.1723                                                                                    &  \\
12 & 0.0833                                                                                    &  \\ \bottomrule
\end{tabular}
\caption{AUC versus SF }
\label{table:auc_sf}
\end{table}
\begin{table}[h]
\centering
\begin{tabular}{@{}lllllll@{}}
\toprule 
                    & Sub1   & Sub2   & Sub3   & Sub4   & Sub5   & Sub6   \\ \midrule
1                   & 0.9936 & 0.9914 & 0.9986 & 0.9929 & 0.8186 & 0.9236 \\
2                   & 0.9979 & 0.9929 & 0.9986 & 0.9993 & 0.8021 & 0.9993 \\
4                   & 0.9957 & 1      & 0.9907 & 0.9993 & 0.7671 & 1      \\
6                   & 0.9943 & 0.9943 & 0.9229 & 0.9907 & 0.7579 & 1      \\
8                   & 0.9700 & 0.9129 & 0.4621 & 0.9008 & 0.4871 & 0.9793 \\
10                  & 0.5336 & 0.2579 & 0.2029 & 0.3669 & 0.2700 & 0.3414 \\
12                  & 0.2500 & 0      & 0.2029 & 0.2684 & 0      & 0.2179 \\ \bottomrule
\end{tabular}
\caption{Accuracy versus SF }
\label{table:accuracy_sf}
\end{table}
\begin{figure}[h]
\begin{center}
   \includegraphics[width=1\linewidth]{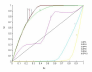}
\end{center}
   \caption{ROC comparisons for different SF’s}
\label{fig:roc_sf}
\end{figure}
\begin{figure}[h]
\begin{center}
   \includegraphics[width=1\linewidth]{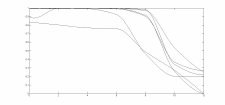}
\end{center}
   \caption{SF vs Accuracy}
\label{fig:accuracy_sf}
\end{figure}
\begin{figure}[h]
\begin{center}
   \includegraphics[width=1\linewidth]{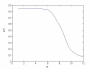}
\end{center}
   \caption{SF vs AUC}
\label{fig:auc}
\end{figure}
\section{Discussions}
The speed versus SF analysis reveals the fact that the speed of operation increases non-linearly with SF. This observation can be explained as follows. With down sampling of images, the number of pixels to be operated on reduces by a factor which is equal to the SF. This reduces the time needed to calculate the integral images. Once the integral images are computed, the time needed for Haar-like feature based detection at any scale and location is constant. Further, the number of sub windows to be searched is also reducing which in turn improves the speed even more.  Hence, such an increase in speed is observed with increase in down sampling SF. The accuracy versus SF shows that the accuracy and AUC remains almost constant upto SF of 6 and then droops down up to an SF of 10 and saturates henceforth. 
\section{Tilted face detection using an affine tranformation}
\begin{figure}[h]
\begin{center}
   \includegraphics[width=1\linewidth]{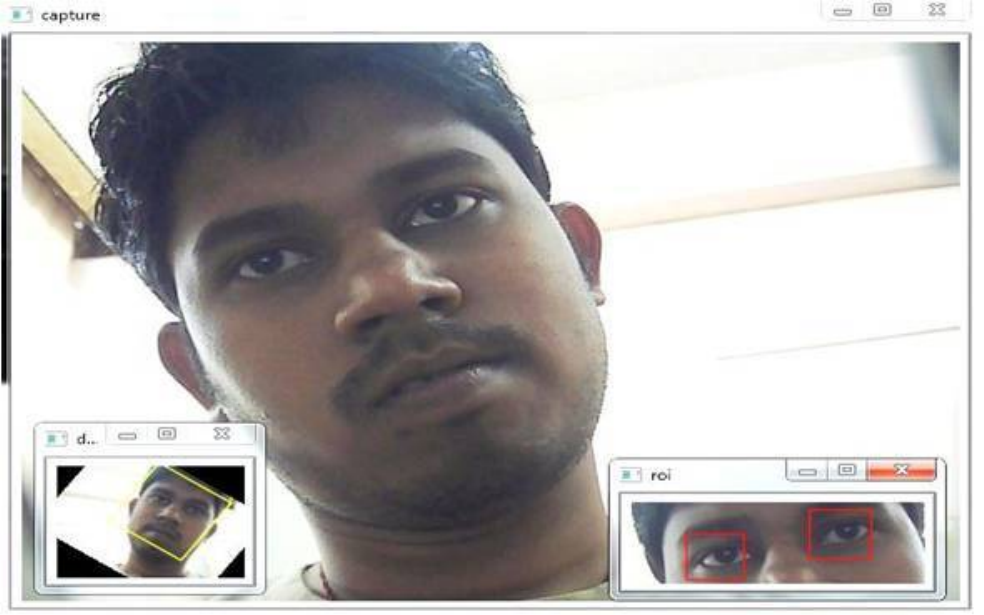}
\end{center}
   \caption{Tilted face detection using an affine transformation}
\label{fig:tilted_face}
\end{figure}

The original Haar cascade technique applied for face detection detects frontal faces only. If there is a moderate amount of tilt of face, it will not be detected. Consequently eyes will also be not detected in such frames. In some applications, tilted face detection is a desired condition and hence an approach for tilted face detection is a must. Several approaches \cite{kong2007multi,hsu2002face,lienhart2002extended,huang2007high} were made for such a purpose. We have adopted an affine transformation based method for the detection of tilted (in plane rotated) images. The rotation matrix can be found for an “n” dimensional image once its size, center and angle of rotation needed are known. This is appended along with down sampling to make a robust face detection algorithm. Fig. \ref{fig:tilted_face} shows tilted face detection using an affine transformation. A detailed description of this algorithm can be found in our earlier works \cite{dasgupta2013board,dasgupta2013vision}. The algorithms hasbeen tested in KGP-NIR fac database \cite{happy2012video}.
\section{Conclusion}
This paper gives an analysis of the speed versus SF and accuracy versus SF. It is observed from the experiment and results that by down sampling up to an SF of 6, there is appreciable amount of increase in speed without much loss of accuracy which improves its real time performance. The method described in this paper can be used in applications to detect eyes along with face. By combining the use of affine transformation to detect tilted faces, the algorithm is made more robust with excellent real-time performance.
\section{Acknowledgements}
The authors would like to thank Mr. S. L. Happy who manually marked the image to find out the ground truth.
\ifCLASSOPTIONcaptionsoff
  \newpage
\fi
\bibliographystyle{IEEEtran}

\bibliography{refs}

\begin{thebibliography}{10}
\providecommand{\url}[1]{#1}
\csname url@samestyle\endcsname
\providecommand{\newblock}{\relax}
\providecommand{\bibinfo}[2]{#2}
\providecommand{\BIBentrySTDinterwordspacing}{\spaceskip=0pt\relax}
\providecommand{\BIBentryALTinterwordstretchfactor}{4}
\providecommand{\BIBentryALTinterwordspacing}{\spaceskip=\fontdimen2\font plus
\BIBentryALTinterwordstretchfactor\fontdimen3\font minus
  \fontdimen4\font\relax}
\providecommand{\BIBforeignlanguage}[2]{{%
\expandafter\ifx\csname l@#1\endcsname\relax
\typeout{** WARNING: IEEEtran.bst: No hyphenation pattern has been}%
\typeout{** loaded for the language `#1'. Using the pattern for}%
\typeout{** the default language instead.}%
\else
\language=\csname l@#1\endcsname
\fi
#2}}
\providecommand{\BIBdecl}{\relax}
\BIBdecl

\bibitem{viola2001rapid}
P.~Viola and M.~Jones, ``Rapid object detection using a boosted cascade of
  simple features,'' in \emph{Computer Vision and Pattern Recognition, 2001.
  CVPR 2001. Proceedings of the 2001 IEEE Computer Society Conference on},
  vol.~1.\hskip 1em plus 0.5em minus 0.4em\relax IEEE, 2001, pp. I--511.

\bibitem{rowley1998neural}
H.~A. Rowley, S.~Baluja, and T.~Kanade, ``Neural network-based face
  detection,'' \emph{Pattern Analysis and Machine Intelligence, IEEE
  Transactions on}, vol.~20, no.~1, pp. 23--38, 1998.

\bibitem{yang1994human}
G.~Yang and T.~S. Huang, ``Human face detection in a complex background,''
  \emph{Pattern recognition}, vol.~27, no.~1, pp. 53--63, 1994.

\bibitem{yang1996real}
J.~Yang and A.~Waibel, ``A real-time face tracker,'' in \emph{Applications of
  Computer Vision, 1996. WACV'96., Proceedings 3rd IEEE Workshop on}.\hskip 1em
  plus 0.5em minus 0.4em\relax IEEE, 1996, pp. 142--147.

\bibitem{yang2000face}
M.-H. Yang, N.~Abuja, and D.~Kriegman, ``Face detection using mixtures of
  linear subspaces,'' in \emph{Automatic Face and Gesture Recognition, 2000.
  Proceedings. Fourth IEEE International Conference on}.\hskip 1em plus 0.5em
  minus 0.4em\relax IEEE, 2000, pp. 70--76.

\bibitem{yow1997feature}
K.~C. Yow and R.~Cipolla, ``Feature-based human face detection,'' \emph{Image
  and vision computing}, vol.~15, no.~9, pp. 713--735, 1997.

\bibitem{zhu2000fast}
Y.~Zhu, S.~Schwartz, and M.~Orchard, ``Fast face detection using subspace
  discriminant wavelet features,'' in \emph{Computer Vision and Pattern
  Recognition, 2000. Proceedings. IEEE Conference on}, vol.~1.\hskip 1em plus
  0.5em minus 0.4em\relax IEEE, 2000, pp. 636--642.

\bibitem{jones2001rapid}
P.~Jones, P.~Viola, and M.~Jones, ``Rapid object detection using a boosted
  cascade of simple features,'' in \emph{University of Rochester. Charles
  Rich}.\hskip 1em plus 0.5em minus 0.4em\relax Citeseer, 2001.

\bibitem{ben2006fuzzy}
M.~Ben~Hmida and Y.~Ben~Jemaa, ``Fuzzy classification, image segmentation and
  shape analysis for human face detection,'' in \emph{Electronics, Circuits and
  Systems, 2006. ICECS'06. 13th IEEE International Conference on}.\hskip 1em
  plus 0.5em minus 0.4em\relax IEEE, 2006, pp. 640--643.

\bibitem{han2000fast}
C.-C. Han, H.-Y.~M. Liao, G.-J. Yu, and L.-H. Chen, ``Fast face detection via
  morphology-based pre-processing,'' \emph{Pattern Recognition}, vol.~33,
  no.~10, pp. 1701--1712, 2000.

\bibitem{hjelmaas2001face}
E.~Hjelm{\aa}s and B.~K. Low, ``Face detection: A survey,'' \emph{Computer
  vision and image understanding}, vol.~83, no.~3, pp. 236--274, 2001.

\bibitem{kong2007multi}
W.-z. Kong and S.-a. Zhu, ``Multi-face detection based on downsampling and
  modified subtractive clustering for color images,'' \emph{Journal of Zhejiang
  University SCIENCE A}, vol.~8, no.~1, pp. 72--78, 2007.

\bibitem{hsu2002face}
R.-L. Hsu, M.~Abdel-Mottaleb, and A.~K. Jain, ``Face detection in color
  images,'' \emph{Pattern Analysis and Machine Intelligence, IEEE Transactions
  on}, vol.~24, no.~5, pp. 696--706, 2002.

\bibitem{lienhart2002extended}
R.~Lienhart and J.~Maydt, ``An extended set of haar-like features for rapid
  object detection,'' in \emph{Image Processing. 2002. Proceedings. 2002
  International Conference on}, vol.~1.\hskip 1em plus 0.5em minus 0.4em\relax
  IEEE, 2002, pp. I--900.

\bibitem{huang2007high}
C.~Huang, H.~Ai, Y.~Li, and S.~Lao, ``High-performance rotation invariant
  multiview face detection,'' \emph{Pattern Analysis and Machine Intelligence,
  IEEE Transactions on}, vol.~29, no.~4, pp. 671--686, 2007.

\bibitem{mohanty2005designing}
P.~K. Mohanty, S.~Sarkar, and R.~Kasturi, ``Designing affine transformations
  based face recognition algorithms,'' in \emph{Proc. IEEE Workshop Face
  Recognition Grand Challenge}, 2005.

\bibitem{fawcett2004roc}
T.~Fawcett, ``Roc graphs: Notes and practical considerations for researchers,''
  \emph{Machine learning}, vol.~31, pp. 1--38, 2004.

\bibitem{dasgupta2013board}
A.~Dasgupta, A.~George, S.~Happy, A.~Routray, and T.~Shanker, ``An on-board
  vision based system for drowsiness detection in automotive drivers,''
  \emph{International Journal of Advances in Engineering Sciences and Applied
  Mathematics}, vol.~5, no. 2-3, pp. 94--103, 2013.

\bibitem{dasgupta2013vision}
A.~Dasgupta, A.~George, S.~Happy, and A.~Routray, ``A vision-based system for
  monitoring the loss of attention in automotive drivers,'' \emph{Intelligent
  Transportation Systems, IEEE Transactions on}, vol.~14, no.~4, pp.
  1825--1838, 2013.

\bibitem{happy2012video}
S.~Happy, A.~Dasgupta, A.~George, and A.~Routray, ``A video database of human
  faces under near infra-red illumination for human computer interaction
  applications,'' in \emph{Intelligent Human Computer Interaction (IHCI), 2012
  4th International Conference on}.\hskip 1em plus 0.5em minus 0.4em\relax
  IEEE, 2012, pp. 1--4.

\end{thebibliography}
\end{document}